\documentclass{article}

\usepackage[preprint]{nips_2018}

\usepackage[utf8]{inputenc} 
\usepackage[T1]{fontenc}    
\usepackage{hyperref}       
\usepackage{url}            
\usepackage{booktabs}       
\usepackage{amsfonts}       
\usepackage{nicefrac}       
\usepackage{microtype}      
\usepackage{graphicx}
\usepackage{amsmath,amssymb}
\usepackage{amsfonts}       
\usepackage{color}
\usepackage{transparent}
\usepackage{natbib}

\usepackage{amsmath}
\usepackage{enumerate}

\usepackage{nicefrac}       

\usepackage{placeins}

\title{Dilated DenseNets for Relational Reasoning}

\author{
  Antreas Antoniou\\
  School of Informatics\\
  University of Edinburgh\\
  \texttt{\scriptsize{a.antoniou@sms.ac.uk}} \\
  \And
   Agnieszka Słowik\\
  School of Informatics\\
  University of Edinburgh\\
  \texttt{\scriptsize{s1778554@sms.ed.ac.uk}} \\
    \AND
   Elliot J. Crowley\\
  School of Informatics\\
  University of Edinburgh\\
  \texttt{\scriptsize{elliot.j.crowley@ed.ac.uk}} \\
  \And
  Amos Storkey\\
  School of Informatics\\
  University of Edinburgh\\
  \texttt{\scriptsize{a.storkey@ed.ac.uk}} \\
}

\begin{document}

\maketitle

\begin{abstract}
Despite their impressive performance in many tasks, deep neural networks often struggle at relational reasoning. This has recently been remedied with the introduction of a plug-in relational module that considers relations between pairs of objects. Unfortunately, this is combinatorially expensive. In this extended abstract, we show that a DenseNet incorporating dilated convolutions excels at relational reasoning on the Sort-of-CLEVR dataset, allowing us to forgo this relational module and its associated expense.

\end{abstract}

\section{Introduction}
\label{sec:intro}

Recent advances in evolution~\citep{darwin1909origin} have allowed humans to excel at image classification (e.g.\ identifying a furry creature as a Shih Tzu) and relational reasoning (e.g.\ noticing that Tom Cruise is much shorter than Dwayne ``The Rock'' Johnson) among other things. Deep Neural Networks similarly excel at image classification but appear to fall short on relational reasoning tasks~\citep{clever}.

In~\cite{rl2017}, the authors present a simple plug-in module that can be appended to existing network architectures to form a {\it relation network}. These achieve state-of-the-art performance on various relational reasoning tasks. They postulate that the inclusion of this flexible module allows the convolutional parts of the network to focus on processing local spatial information. Unfortunately, this module is combinatorially expensive as it performs operations on pairs of features that are outputted from a convolutional neural network (CNN).

In this work, we show that this module can be avoided by simply incorporating {\it dilated} convolutions~\citep{dilated2016} into a powerful CNN architecture, such as a DenseNet~\citep{huang2017densely}. These {\it Dilated DenseNets} exhibit comparable performance to relation networks on the Sort-of-CLEVR~\citep{rl2017} dataset without the need for a relational reasoning module.

Our contributions are as follows:

\begin{itemize}
\item We introduce a modification for DenseNets which enables them to achieve multi-scale feature learning and relational reasoning.
\item We empirically show that this Dilated DenseNet can achieve strong relational reasoning results without the need for an explicit relational module.
\item We showcase that usage of an explicit relational module is redundant; we don't have to incur the computational costs of having to train a model with this module.
\end{itemize}

\section{Background}

\paragraph{SORT-of-CLEVR} In this work we train our models to solve SORT-of-CLEVR. This dataset --- introduced in~\cite{rl2017} --- consists of 10,000 images each containing squares and circles in random locations. These shapes can be one of six colours. Given an image, the task is to answer a question which can be relational (e.g.\ {\it What shape is the object closest to the blue object?}) or non-relational (e.g.\ {\it What shape is the red object?}). There are 10 relational and 10 non-relational questions in total.

\paragraph{Relation networks} We benchmark against relation networks~\citep{rl2017}. These consist of a CNN, an LSTM~\citep{Hochreiter:1997:LSM:1246443.1246450}, and a relational module. The image in each image-question pair is passed through the CNN to produce a set of {\it objects}: one for each 2D spatial location of the CNN output, with a corresponding feature given by the values of each channel at that location. The question is fed into the LSTM, and its output is appended to each vector of object-pair features. Finally, the relational module takes all these vectors, puts them through an MLP and takes the sum to produce an answer. All weights are learnt in an end-to-end manner.

\begin{figure}[h]
        \centering
        \includegraphics[width=.8\textwidth]{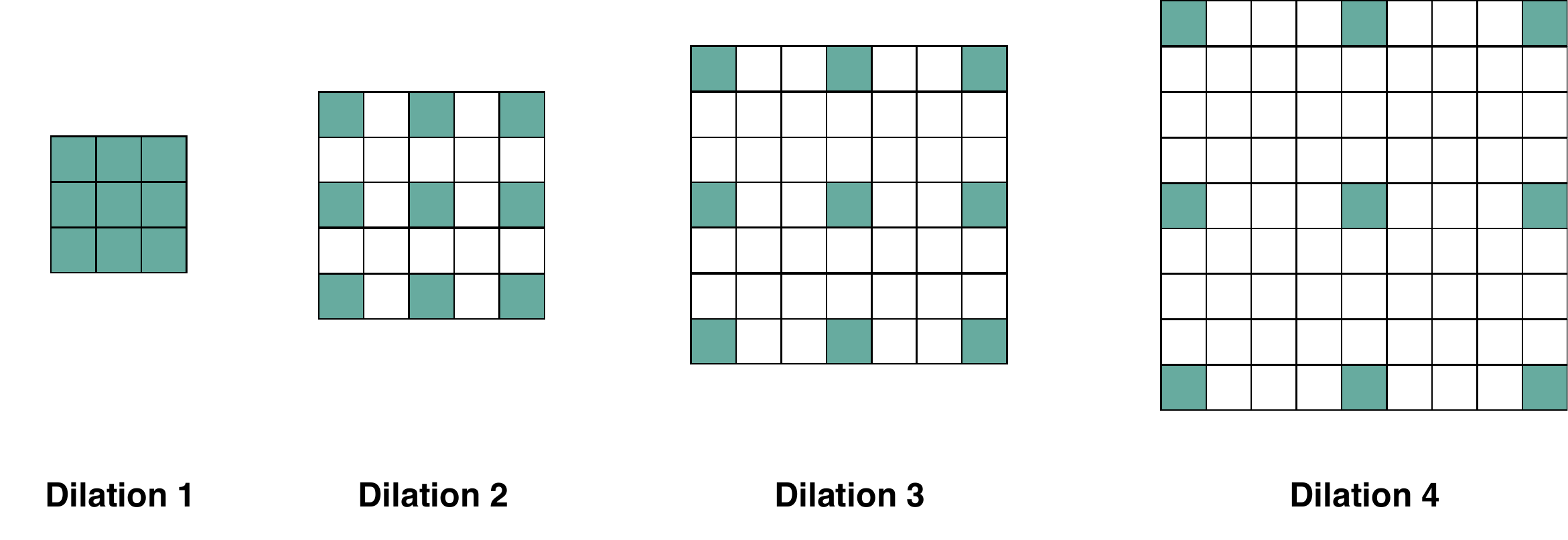}
        \caption{An illustration showing the effect of increasing the dilation of a $3 \times 3$ filter. For a given image patch, each coloured square corresponds to the locations at which the filter is placed. Notice that this has the effect of increasing the receptive field of the filter as dilation is incremented.}
\label{fig:dilations}
\end{figure}

\paragraph{Dilated Convolutions}  Dilated convolutions are effectively convolutions with expanded receptive fields. For example, in a standard convolution with $3\times3$ kernels and a stride of 1, the filters scan the image in $3\times3$ regions of adjacent pixels. This filter has {\it dilation} 1: centre-to-centre, there is a 1-pixel distance between each filtered pixel and its nearest neighbour. Now, consider the case where there is a 2-pixel distance: the filter is only applied to pixels that are in both odd-numbered rows and columns in each $5\times 5$ region. This is dilation 2. For dilation 3, the filter is applied to only pixels in every third row and column in each $7\times 7$ region, and so on. These are illustrated in Figure~\ref{fig:dilations}. These dilations allow a model to learn higher order abstractions without the need for dimensionality reduction. They are frequently used in segmentation networks~\citep{dilated2016,Romera2017a,Romera2017b}, but can also be used for model compression~\citep{moonshine}, and audio generation~\citep{2016arXiv160903499V} among other things.

\paragraph{DenseNets} DenseNets~\citep{huang2017densely} are a powerful family of architectures that encourage feature reuse. They consist of a series of repeating convolutional blocks; the outputs of earlier blocks are concatenated and form the input to later blocks. Typically this block contains either (i) a single $3\times 3$ convolution, usually referred to as a \emph{basic} block, or (ii) a bottleneck $1\times 1$ convolution followed by a $3\times 3$ convolution (a \emph{bottleneck} block). Each block has the same number of output channels, this is the network's \emph{growth rate} --- it controls the rate at which the network expands. A standard DenseNet comprises of 3 \emph{dense stages}, each consisting of a number of blocks. After each stage, a transition layer is applied, which carries out spatial dimensionality reduction using an average pooling layer and channel-wise dimensionality reduction using a bottleneck $1 \times 1$ convolution.


\section{Model}
\label{sec:model}
Our proposed model consists of a standard DenseNet utilising basic blocks, which is modified by changing the per-dense-stage convolutions to follow an exponentially increasing dilation rate assignment. The dilation rate assignment scheme can be expressed as:
\[r_{ij} = 2^i \]
Where $r_{ij}$ is the dilation rate of the convolution at the $i$th block (for $i=0,1,2...$) of the $j$th dense stage. For example, if each stage consists of 5 blocks, each consisting of 1 convolutional layer, then those layers will have dilation rates of 1, 2, 4, 8, 16 respectively. 

Specifically, we use a 16-layer DenseNet with a growth rate of 32. A compression factor of 1 is used for the transition layers.

\section{Experiments}

In this section we run experiments on the SORT-of-CLEVR dataset to test the relational, and non-relational accuracy of different models, including our Dilated DenseNet.

We first reproduce the original relation network from~\cite{rl2017}. It consists of a standard 4 layer convolutional network with batch normalisation and ReLU activation functions, followed by a relational module, and a final softmax layer. This is denoted as CNN + RN in Table~\ref{tab:results}. We also train the same network where the relational module has been replaced by a 2-layer MLP (CNN + MLP). We can see from Table~\ref{tab:results} that while both networks succeed at non-relational reasoning, the network with a relational module attains a significantly higher relational accuracy than the one without it (87.7\% vs. 65.4\%). It is practically a required component in this case.

We compare these to our network consisting of a Dilated DenseNet and a 2-layer MLP (Dilated DenseNet + MLP). Notice that it attains a relational accuracy of 83.7\%,
a few points shy of that achieved by the relation network --- and a gigantic step up from CNN + MLP --- without any need for a relational reasoning module.

Finally, we train a DenseNet without dilations (DenseNet + MLP) to observe where the dilations are required for relational reasoning. It transpires that they are; not only does this network fail to perform as well as its dilated counterpart --- it is only marginally better than CNN + MLP at relational reasoning --- it also exhibits very high deviation in performance across three independent training runs, particularly on non-relational questions.

\paragraph{Implementation Details}

All networks are trained using the Adam optimiser~\citep{adam} with an initial learning rate of $10^{-3}$ and default momenta. This is cosine annealed to $10^{-5}$ over 250 epochs. For the DenseNets we use a weight decay of $4\times 10^{-5}$ and a dropout rate of 0.2 in every layer. For the standard CNNs we used no weight decay and a dropout rate of 0.5 between the 2 layers of the MLP. After each epoch, we evaluate our model on a validation set. The model that performs best on the validation set across a training run is then evaluated on the test set. We perform three independent runs per network with different seeds for the model initialisation and the data provider.

\begin{table}[t]
\caption{Results on the test set of SORT-of-CLEVR: We display accuracy on the non-relational questions, accuracy on the relational questions, and the combination thereof. Means and standard deviations are given across three independent runs. CNN + MLP is a simple 4 layer neural network with an MLP. CNN + RN is the same neural network with a relational reasoning module. Our proposed network (Dilated DenseNet + MLP)  achieves strong relational reasoning results, very close to those of CNN + RN, without the need for an explicit relational reasoning module. We also compare this to its undilated counterpart (DenseNet + MLP) to show that the dilations are indeed necessary.
}
\begin{tabular}{|l|r|r|r|}
\hline
\textbf{Model}         & \textbf{Non-relational acc.} & \textbf{Relational acc.} & \textbf{Combined acc.} \\ \hline

CNN + MLP              & $98.2 \pm 0.4\%$ & $65.4\pm1.1\%$ & $81.7 \pm 0.5\%$                     \\ \hline
CNN + RN               & $99.7 \pm 0.1\%$ & $87.7\pm5.1\%$ &  $93.6 \pm 2.6\%$                    \\ \hline
DenseNet + MLP & $87.3 \pm 15.7\%$ & $66.7\pm 3.5\%$ & $76.9 \pm 7.0\%$                \\ \hline
Dilated DenseNet + MLP  & $99.2 \pm 0.1\%$ & $83.7\pm2.3\%$& $91.3 \pm 1.2\%$               \\ \hline
\end{tabular}
\label{tab:results}
\end{table}

\section{Conclusion}

Relational reasoning is an important task, one in which neural networks were believed to fail at without the addition of an expensive tailored module. In this work we have demonstrated that this is not the case. By taking a powerful network architecture and incorporating dilations we are able to forgo this module, and its associated costs. Future work could entail applying our network to further relational reasoning tasks~\citep{2015arXiv150205698W,clever}

\bibliography{mybib}
\bibliographystyle{icml2018}

\end{document}